\documentclass[cameraready]{Interspeech}

\title{Regularized Entropy Information Adaptation with Temporal-Awareness Networks for Simultaneous Speech Translation}

\author[equalcontribution]{Joseph}{Liu}
\author[equalcontribution]{Nameer}{Hirschkind}
\author{Xiao}{Yu}
\author{Mahesh Kumar}{Nandwana}

\address{
    Roblox
}

\email{\texttt{{\{josephliu,nhirschkind,xyu,mnandwana\}@roblox.com}}}


\usepackage{comment}
\usepackage{subcaption}
\usepackage{multirow}
\usepackage{balance}

\begin{document}

\maketitle

\begin{abstract}

Simultaneous Speech Translation (SimulST) requires balancing high translation quality with low latency. Recent work introduced REINA, a method that trains a \textsc{Read}/\textsc{Write} policy based on estimating the information gain of \textsc{read}ing more audio. However, we find that information-based policies often lack temporal context, leading the policy to bias itself toward reading most of the audio before starting to write. 
We improve REINA using two distinct strategies: a supervised alignment network (REINA-SAN) and a timestep-augmented network (REINA-TAN). Our results demonstrate that while both methods significantly outperform the baseline and resolve stability issues, REINA-TAN provides a slightly superior Pareto frontier for streaming efficiency, whereas REINA-SAN offers more robustness against 'read loops'. Applied to Whisper, both methods improve the pareto frontier of streaming efficiency as measured by Normalized Streaming Efficiency (NoSE) scores up to 7.1\% over existing competitive baselines.


\end{abstract}

\section{Introduction}

Simultaneous Speech Translation (SimulST) aims to generate target-language text in real-time while source speech is still unfolding. Unlike offline Speech-to-Text Translation (S2TT), SimulST systems must decide at each step whether to wait for more audio (\textsc{Read}) or emit the next target token (\textsc{Write}), inducing a fundamental trade-off between translation quality and latency. Excessive waiting improves translation quality but degrades responsiveness, while overly aggressive generation harms translation quality.

Recent work has explored adaptive \textsc{Read}/\textsc{Write} policies that dynamically navigate this trade-off. Among these, Regularized Entropy Information Adaptation (REINA)~\cite{reina_hirschkind2025reina} formulates the decision as an information-theoretic problem: the system should wait only when future audio is expected to significantly reduce uncertainty about the next token. This formulation is attractive because it decouples the streaming policy from the underlying translation model, enabling lightweight adaptation of strong offline models.

However, in practice, REINA suffers from an apparent deficiency in temporal awareness, limiting its reliability across models and datasets. We observe the policy can enter a degenerate behavior in which it repeatedly predicts \textsc{Read} despite consuming additional audio, delaying all output until the end of the utterance. We refer to this behavior as a \emph{read loop}. While not exceedingly common, the presence of read loops suggests REINA models may not have a strong internal representation of time.

In this work, we address this issue via two methods. First, we draw on existing works such as CLASI~\cite{clasi_cheng2024towards}, Hibiki~\cite{hibiki_labiausse2025high} and Hibiki-Zero~\cite{hibikizero_labiausse2026simultaneous} and add a weak monotonic supervision signal derived from LLM-generated alignments. Second, we try a simple augmentation to REINA, adding an explicit encoding of the consumed audio duration to provide temporal awareness to the policy. We find REINA with a Timestep Augmented Network (REINA-TAN) and REINA with a Supervised Alignment Network (REINA-SAN) both consistently improve the latency-quality tradeoff, with REINA-TAN taking a slight lead.

We provide the first comprehensive evaluation of REINA-style policies on a large, open-source speech translation model across multiple benchmarks and language directions. Our results show that REINA-TAN achieves state-of-the-art streaming efficiency as measured by Normalized Streaming Efficiency (NoSE)~\cite{reina_hirschkind2025reina}, validating the scalability of information-based policies to modern foundation models. Our main contributions are:
\begin{itemize}
   \item An empirical analysis of information-theoretic policies, demonstrating that information gain alone without explicit temporal grounding leads to sub-optimal latency-quality trade-offs in large-scale models.

   \item REINA-TAN, a novel architectural enhancement that injects continuous temporal embeddings into the policy head, improving the streaming pareto frontier.

   \item REINA-SAN, a supervised alignment method that uses LLM-generated, monotonically aligned data to learn the optimal policy in a supervised manner.

   \item We demonstrate that REINA-TAN and REINA-SAN scales effectively to foundation-scale models (Whisper Large V3), achieving state-of-the-art results in Normalized Streaming Efficiency (NoSE) across multiple benchmarks and language pairs.
\end{itemize}

\section{Related Work}

The transition from offline Speech-to-Text Translation (S2TT) to SimulST necessitates a read/write policy to balance quality and latency. Fixed policies, such as wait-$k$~\cite{waitk-ma-etal-2019-stacl, ma2020monotonic}, are simple to implement but lack the flexibility to handle varying speech rates and word reorderings across languages. Adaptive policies address this by dynamically determining when sufficient context has been processed.

\textbf{Architecture-Integrated Policies.} Many adaptive approaches integrate the policy directly into the translation model's architecture. Monotonic Multihead Attention (MMA)~\cite{emma, arivazhagan2019monotonic} and neural transducers~\cite{graves2012sequence, xue2022large} learn alignment jointly with translation. Other methods use reinforcement learning~\cite{seedliveinterpret_cheng2025seed} to find the optimal policy. While expressive, these methods require expensive training from scratch or extensive fine-tuning, often suffering from numerical instability and convergence difficulties~\cite{reina_hirschkind2025reina,seedliveinterpret_cheng2025seed}. This makes them unsuitable for adapting large-scale foundation models like Whisper~\cite{whisper-pmlr-v202-radford23a}, Canary~\cite{canary}, or OWSM~\cite{owsm} without prohibitive computational cost. Moreover, fine-tuning of these backbones for streaming risks catastrophic forgetting of the translation quality.

\textbf{Streaming Foundation Models.} Recent efforts have focused on adapting foundation models for streaming. Common approaches apply heuristic segmentation to frozen models, such as Voice Activity Detection (VAD) or fixed-size block processing~\cite{ren-etal-2020-simulspeech}. Local agreement methods~\cite{liu2020low} allow re-transcribing overlapping chunks. Another class of models, including Hibiki~\cite{hibiki_labiausse2025high} and SASST~\cite{sasst}, fine-tune Large Language Models (LLMs) on artificially aligned data derived from heuristics or teacher models. While effective, these strategies are limited by the quality of the external alignments and do not directly optimize the simultaneous translation performance (latency-quality tradeoff) of the foundation model itself.

\textbf{Decoupled Policy Networks.} To combine the power of foundation models with adaptive streaming, recent works train lightweight policy networks on top of frozen backbones. DiG-SST~\cite{dig_sst_2024} trains a policy to minimize the divergence between the output distributions of partial and full inputs. Building on this, REINA~\cite{reina_hirschkind2025reina} formulates the decision as an information-theoretic problem, estimating the mutual information gain of waiting for future audio. The original work hypothesizes that this formulation should inherently scale to large foundation models, but restricts evaluation to medium-scale encoders. Our work provides the first empirical validation of that claim on large-scale open-source models, while identifying and resolving the policy's deficiency in temporal awareness that arise in this regime.

\section{Preliminaries}
Simultaneous Speech Translation requires a policy to decide between \textsc{Read} (consuming audio frame $a_{t+1}$) and \textsc{Write} (emitting token $s_{n+1}$) at step $(t, n)$. REINA derives this policy from the principle of Information Maximization: the system should wait for more context if and only if doing so significantly reduces uncertainty about the next token.

\textbf{Information Gain.} We quantify the value of waiting as the mutual information gain $\mathcal{F}$ regarding the next token $s_{n+1}$ provided by the full audio context $a_T$ relative to the current partial context $a_t$:
\begin{equation}
    \mathcal{F}(t, n) = I(s_{n+1}; a_T, S_n) - I(s_{n+1}; a_t, S_n)
\end{equation}
Rewriting in terms of conditional entropy, this simplifies to the expected difference in log-likelihoods assigned to the ground truth token by a model with full versus partial access:
\begin{equation}
    \mathcal{F}(t, n) \approx \log P(s_{n+1} | a_T, S_n) - \log P(s_{n+1} | a_t, S_n)
\end{equation}
Intuitively, $\mathcal{F}$ is large when the future audio contains critical information (e.g., a verb at the end of a German sentence) that resolves ambiguity over what $s_{n+1}$ should be.

\textbf{Covariance Maximization.} We train a lightweight policy network $q_\theta$ to estimate this information gain. Rather than regressing directly to the noisy log-probability values, REINA trains $q_\theta$ to maximize the covariance with $\mathcal{F}$. The primary loss function minimizes the product of $q_\theta$ and the batch-normalized information gain:
\begin{equation}
    \mathcal{L}_{cov} = \frac{1}{B} \sum_{i=1}^B q_\theta^{(i)} \cdot \text{BN}\left[ \log P(s\mid a_t) - \log P(s\mid a_T) \right]
\end{equation}
where BN denotes batch normalization. We obtain a policy from $q_\theta$ by simple thresholding: \textsc{Read} if and only if $q_\theta > \alpha$ for a tunable hyper-parameter $\alpha$.

\textbf{Regularization.} To ensure stable streaming behavior, the total objective $\mathcal{L}_{REINA}$ includes auxiliary terms for $L_2$ regularization and a monotonicity constraint $\mathcal{L}_{mono}$, which encourages $q_\theta$ to estimate larger and larger information gain over a token sequence given fixed audio.

\section{Motivation: Analysis of Policy Instability}
\subsection{Analysis of REINA}
\begin{figure}
    \centering
    \includegraphics[width=1\linewidth]{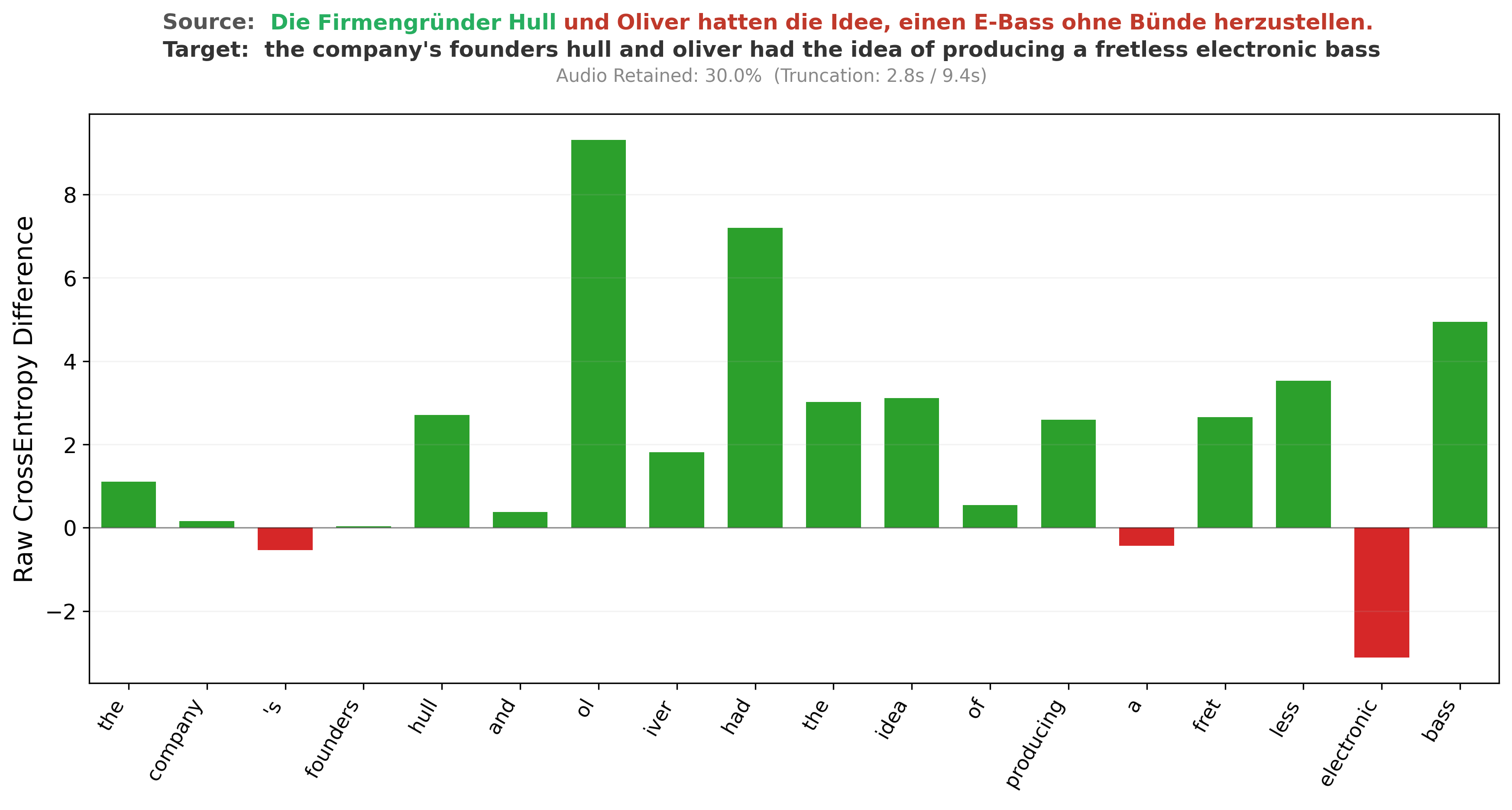}
    \vspace{-8mm}
    \caption{\textbf{Information Gain} Given a fixed \% of source audio, we plot information gain for each label token. There is a clearly visible \textsc{Read}-\textsc{Write} boundary at the token  \textsc{hull}.}
    \label{fig:analysis}
    \vspace{-5mm}
\end{figure}

While the original REINA article~\cite{reina_hirschkind2025reina} provides substantial theoretical justification of the REINA loss, empirical analysis of the loss is limited. As shown in figure \ref{fig:analysis}, we plot the estimated information gain at train time given partial audio to understand visually how the loss works. We observe that in most cases the labels provide a clear boundary of where the model should switch from \textsc{Write} to \textsc{Read} given a fixed truncated audio. This motivates continuing to use the same information gain label. We also experimented with a simple MSE loss on the policy head outputs using the same information gain label and found the resulting policy to underperform the covariance maximization loss in the original paper.
This suggests the ordinal relationship of information gain is more critical for policy stability than its absolute magnitude, which favors the covariance maximization objective.

\begin{figure}[b]
    \centering
    \vspace{-7mm}
    \includegraphics[width=0.8\linewidth]{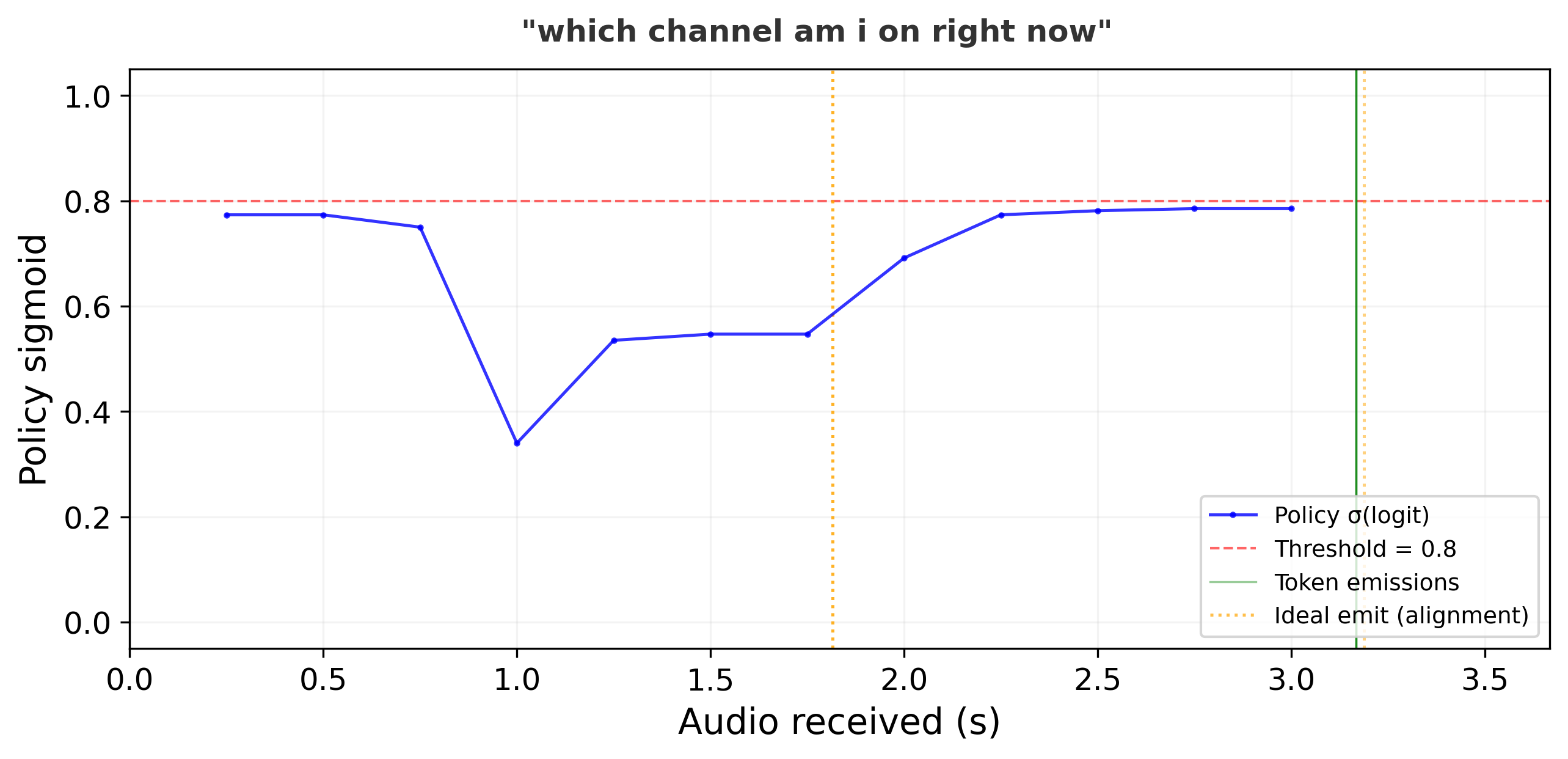}
    \vspace{-4mm}
    \caption{\textbf{Read Loop} Despite ingesting more and more audio over time, the policy continues to predict roughly the same value, corresponding to \textsc{Read} decisions and entering a Read Loop.}
    \label{fig:readloop}
    \vspace{-5mm}
\end{figure}
\subsection{Deficient Temporal Awareness}
Empirical analysis suggests that information-based policies may suffer from temporal drift, which is a state where the policy, lacking an internal clock, fails to increase its emission probability as audio duration grows, as seen in figure \ref{fig:readloop}. This leads to unstable waiting strategies during inference where the system remains in a 'Read' state indefinitely, entering a ``read loop". We hypothesize that explicit temporal grounding is required to stabilize these decisions.
\section{Method}


\begin{figure}[t]

  \vspace{-3mm}
  \centering
  \begin{subfigure}{0.49\columnwidth}
    \centering
    \includegraphics[width=0.9\textwidth]{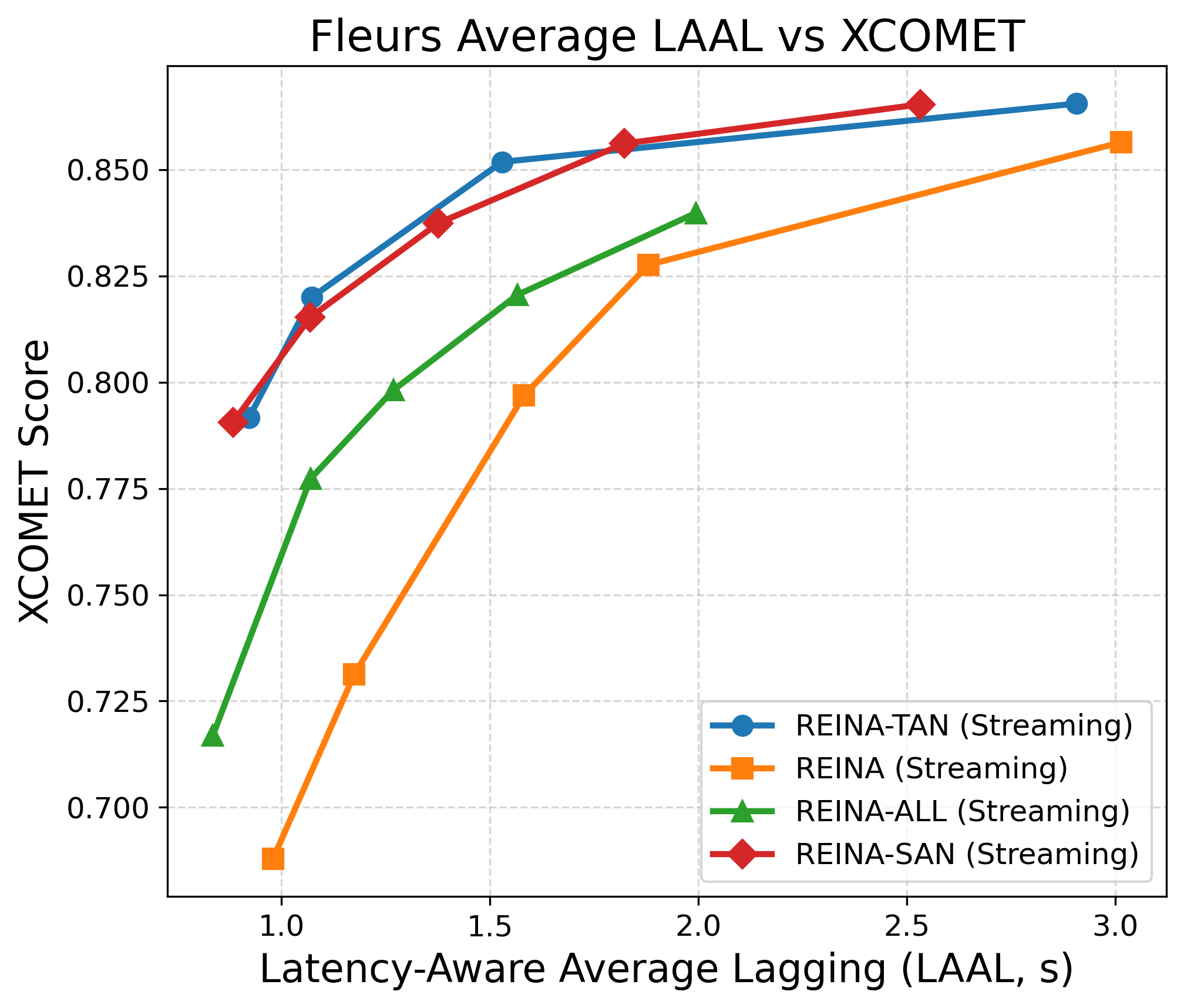}
    \caption{FLEURS (de/fr/es$\rightarrow$en).}
    \label{fig:laal_xcomet_fleurs}
  \end{subfigure}
  \hfill
  \begin{subfigure}{0.49\columnwidth}
    \centering
    \includegraphics[width=0.9\textwidth]{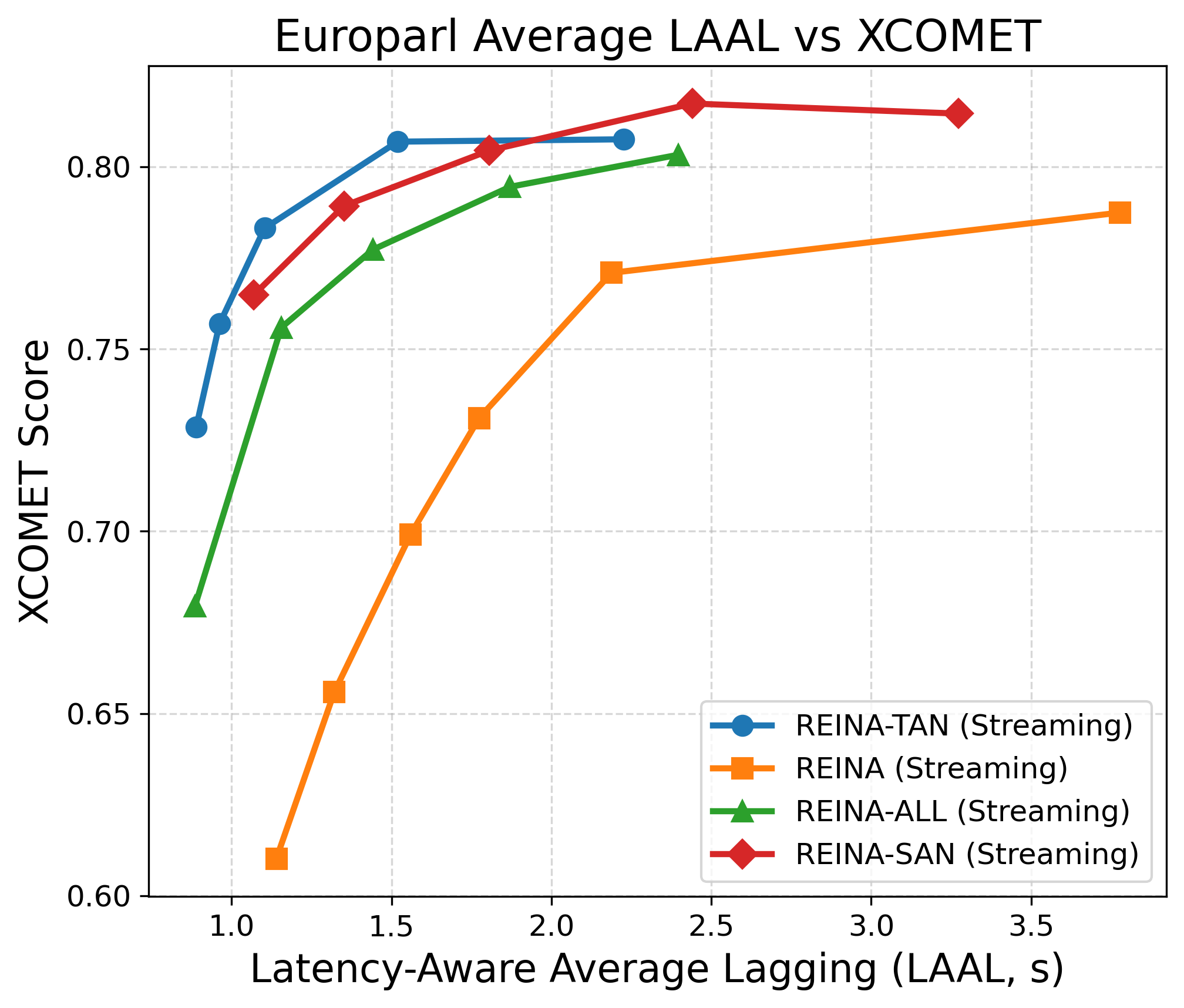}
    \caption{EuroparlST (de/fr/es$\rightarrow$en).}
    \label{fig:laal_xcomet_europarl}
  \end{subfigure}
  \vspace{-3mm}
  \caption{Latency--quality trade-off of Whisper policy ablations.
  We plot XComet-XL versus LAAL by sweeping the policy threshold and averaging over de/fr/es$\rightarrow$en.
  REINA-TAN achieves the strongest Pareto frontier across both datasets.}
  \label{fig:laal_xcomet_combined}
  \vspace{-5mm}
\end{figure}
We experiment with two methods to improve REINA's temporal awareness and streaming efficiency:  (i) A Supervised Alignment Network that leverage LLM-generated alignments, and (ii) a Timestep Augmented Network that augments the policy network with a duration encoding.

\subsection{Supervised-Alignment Network}
We introduce a secondary supervision signal derived from monotonic alignments generated by a large language model (LLM), inspired by EASIST~\cite{easist} and CLASI~\cite{clasi_cheng2024towards}.

\textbf{Generating Alignment Targets.} We first use an LLM to generate a monotonic segmentation of the source transcript and target translation. We then align these text chunks to the source audio using a forced aligner to obtain a set of ``ideal" emission times for tokens of the target text. For each target token $s_n$, we derive an ideal audio boundary time $t^*_n$ by which the token should be emitted.

\textbf{Soft Labeling.} Instead of forcing a hard boundary, we apply a loss using soft labels. We define a target probability $y^*_{align}$ for the policy to output \textsc{Write} (or \textsc{Read} depending on convention) based on the current audio duration $t_{audio}$:
\begin{equation}
    y^*_{align}(n, t_{audio}) = \sigma\left(\frac{t_{audio} - t^*_n}{\tau} \right)
\end{equation}
where $\sigma$ is the sigmoid function and $\tau$ is a temperature parameter controlling the sharpness of the boundary. When $t_{audio} > t^*_n$, the target approaches 1 (Emit); when $t_{audio} \ll t^*_n$, it approaches 0 (Wait).

\textbf{Combined Objective.} We effectively treat this as a multi-task learning problem. The total loss combines the uncertainty-based REINA loss with an alignment loss:
\begin{equation}
    \mathcal{L}_{align} = \text{BCE}(q_\theta, y^*_{align})
\end{equation}
We then combine the losses in a hybrid objective:
\begin{equation}
\mathcal{L}_{REINA-SAN} = \mathcal{L}_{REINA} + \lambda_{align} \mathcal{L}_{align}
\end{equation}
This hybrid objective allows the policy to follow the information gain signal for fine-grained decisions while being constrained by the robust monotonic alignment derived from the LLM. Note that we treat $p_\theta$ as a logit here, which is not the case in $\mathcal{L}_{REINA}$. However, due to the L2 regularization term in REINA,  $q_\theta$ already tends to cluster around 0. Empirically we find this combined loss converges stably.

\subsection{Timestep-Augmented Network}
To address the REINA policy network's lack of temporal awareness, we inject an explicit audio duration encoding. We employ a fixed sinusoidal embedding similar to the positional encoding in Transformers~\cite{vaswani2017attention}, but applied to continuous time values and with a different periodicity. We use a base of 100 to increase the resolution of the encoding across the typical duration of a spoken utterance (5-30 seconds). Let $t_{audio}$ be the duration of audio consumed in seconds. We map this scalar to a vector $e_{time} \in \mathbb{R}^d$:
\begin{equation}
    e_{time}^{(2i)} = \sin\left(\frac{t_{audio}}{100^{2i/d}}\right), \quad e_{time}^{(2i+1)} = \cos\left(\frac{t_{audio}}{100^{2i/d}}\right)
\end{equation}
This embedding is broadcast and added to the hidden states of the decoder $H_{dec}$ as input before being fed into the policy network: $H_{policy} = H_{dec} + e_{time}$.
This simple modification provides the policy with a robust ``clock", giving it a strong signal that more audio has been ingested, warranting a higher likelihood of \textsc{Write}-ing. 

\section{Experiments}
We experiment with augmenting REINA with weak supervision from LLM alignments as well as an audio duration encoding. We ultimately show that the latter (REINA-TAN) performs best, improving upon the previous state of the art for simulST.


\begin{figure}[t]
    \vspace{-6mm}
    \centering
    \includegraphics[width=0.8\linewidth]{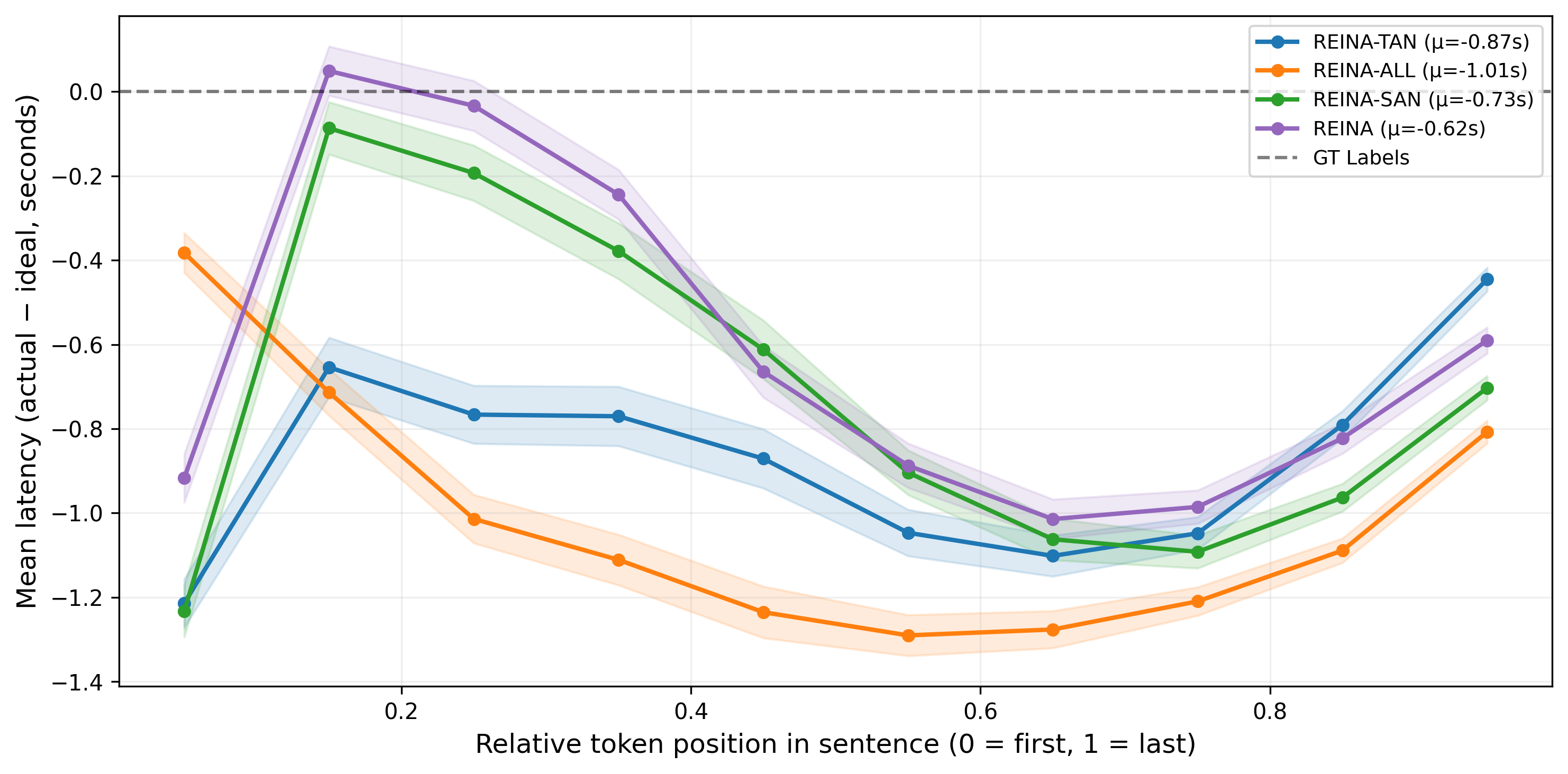}
    \vspace{-3mm}
    \caption{Mean per-token latency (actual emission time minus monotonic alignment time boundaries) as a function of relative token position within the sentence, with 95\% confidence intervals for 2000 samples on the dev split of CVSS-C.}
    \label{fig:latency-vs-position}
    \vspace{-6mm}
\end{figure}

\subsection{Experimental Setup}

\textbf{Model Variants.} We build on top of a strong offline speech translation model: Whisper Large V3~\cite{whisper-pmlr-v202-radford23a}. We freeze all parameters, augment it with a lightweight transformer REINA policy module, and train the policy module using several variants of our proposed loss function. To test the effectiveness of timestep augmented networks and supervised alignment networks, we experiment with four policy variants: REINA only (REINA), REINA with the timestep augmented network (REINA-TAN), REINA with $\mathcal{L}_{align}$ (REINA-SAN), and REINA with timestep augmented networks and $\mathcal{L}_{align}$ (REINA-ALL).

\begin{table}[h]
\centering
\fontsize{5.5}{6}\selectfont 
\setlength{\tabcolsep}{2.5pt}
\caption{Streaming evaluation results on FLEURS and Europarl benchmarks. Operating points are selected to represent low and medium latency regimes across languages.}
\vspace{-2mm}
\label{tab:streaming_results_curated}
\begin{tabular}{l ccc c ccc}
\toprule
 & \multicolumn{3}{c}{\textbf{FLEURS}} & & \multicolumn{3}{c}{\textbf{Europarl}} \\
 \cmidrule{2-4} \cmidrule{6-8}
\textbf{Model} & \textbf{LAAL(s)} & \textbf{BLEU} & \textbf{X-COMET} & & \textbf{LAAL(s)} & \textbf{BLEU} & \textbf{X-COMET} \\
\midrule
\multicolumn{8}{c}{\textit{De$\rightarrow$En}} \\
\midrule
REINA & 1.33 & 25.2 & .696 & & 1.40 & 16.3 & .608 \\
REINA-All & 1.21 & 26.9 & .772 & & 1.26 & 21.8 & .760 \\
REINA-SAN & 1.03 & 28.6 & .794 & & 1.09 & 21.4 & .755 \\
REINA-TAN & 1.03 & 27.9 & .775 & & 0.98 & 20.5 & .721 \\
\cmidrule{1-8}
REINA & 1.82 & 29.7 & .788 & & 1.80 & 19.8 & .698 \\
REINA-All & 1.79 & 30.2 & .824 & & 1.94 & 23.6 & .806 \\
REINA-SAN & 1.99 & 33.0 & .866 & & 2.05 & 23.8 & .825 \\
REINA-TAN & 1.79 & 33.0 & .860 & & 1.72 & 23.9 & .816 \\

\midrule
\multicolumn{8}{c}{\textit{Es$\rightarrow$En}} \\
\midrule
REINA & 1.11 & 18.8 & .792 & & 1.12 & 26.4 & .735 \\
REINA-All & 0.97 & 18.5 & .801 & & 1.05 & 27.5 & .758 \\
REINA-SAN & 1.01 & 19.8 & .828 & & 1.10 & 28.7 & .779 \\
REINA-TAN & 0.99 & 20.3 & .835 & & 1.01 & 29.0 & .789 \\
\cmidrule{1-8}
REINA & 1.83 & 21.4 & .844 & & 2.37 & 30.0 & .796 \\
REINA-All & 1.81 & 21.1 & .848 & & 1.75 & 29.6 & .792 \\
REINA-SAN & 1.72 & 21.4 & .865 & & 2.04 & 30.5 & .810 \\
REINA-TAN & 2.02 & 22.3 & .869 & & 1.98 & 30.2 & .800 \\

\midrule
\multicolumn{8}{c}{\textit{Fr$\rightarrow$En}} \\
\midrule
REINA & 1.06 & 24.8 & .694 & & 1.23 & 20.9 & .607 \\
REINA-All & 1.03 & 26.4 & .753 & & 1.10 & 27.0 & .745 \\
REINA-SAN & 0.94 & 27.9 & .795 & & 1.01 & 27.8 & .764 \\
REINA-TAN & 1.02 & 28.0 & .797 & & 0.99 & 28.1 & .777 \\
\cmidrule{1-8}
REINA & 2.54 & 30.9 & .832 & & 3.29 & 29.6 & .786 \\
REINA-All & 1.91 & 30.1 & .817 & & 2.46 & 29.4 & .789 \\
REINA-SAN & 2.51 & 31.5 & .836 & & 2.61 & 30.2 & .808 \\
REINA-TAN & 2.52 & 31.5 & .848 & & 1.90 & 29.8 & .803 \\
\bottomrule
\end{tabular}
\vspace{-8mm}
\end{table}
\textbf{Data} The models were trained on a mixture of two datasets: a) The fr, de, es $\xrightarrow{}$ en splits of CVSS-C~\cite{cvss} (totalling 563 hours of audio), and b) The fr, de, and es splits of Multilingual LibriSpeech (MLS)~\cite{pratap2020mls} (3327 hours) with transcripts translated into English using Gemma-2-9B~\cite{team2024gemma}. We run Metric-X~\cite{metricx} quality estimation on our generated translations and only retain translations with a score of less than 4.

\textbf{Generating monotonic alignments} To generate ground truth labels for REINA-SAN, we first generate alignments for MLS and CVSS using WhisperX~\cite{whisperx_Bain2023WhisperXTS}, and then use Qwen3-32B~\cite{qwen3} with the instruction to generate monotonic alignments between source and target text. The instructions specifically mention to pair the smallest possible semantically relevant chunks from source and target text without reordering the translation. These chunks are systematically validated to not mutate both source and target labels and invalid samples would go through a second round. Samples that were not aligned successfully remained in the dataset for training, just without being used for the alignment loss. 

\textbf{Evaluation} We evaluate Whisper on the fr, de, es $\xrightarrow{}$ en splits of FLEURS~\cite{fleurs}, and the same splits from EuroparlST~\cite{europarl}. To align with established literature, we run Silero VAD on FLEURS to trim the silence at the beginning and end of the source audios, but do not apply VAD to Europarl.

\textbf{Metrics} We evaluate latency using the common metric Length Adaptive Average Lagging (LAAL)~\cite{laal}. We evaluate translation quality using the common BLEU metric, implemented with the SacreBLEU package~\cite{sacrebleu}. 
However, since neural translation quality metrics have been shown to correlate better with human judgment compared to lexical ones, we also use the neural metric XComet-XL~\cite{xcomet} for translation quality. 
We adopt Normalized Streaming Efficiency (NoSE)~\cite{reina_hirschkind2025reina} to measure the efficiency of conversion from a non-streaming speech translation model to a SimulST model. Finally, we measure the percentage of samples on which the model \textsc{Read}s until all audio has been ingested before \textsc{Write}ing and call this our ``Read Loop \%".

\textbf{Baselines} In addition to our own ablations, we compare to Seamless~\cite{seamless_barrault2023seamlessm4t}, the state of the art in low latency SimulST.

\textbf{Training} We train the models for 4 epochs on 8 A100-80G gpus using an inverse cosine scheduler on top of an AdamW optimizer with 1e-4 weight decay and 10k warmup steps, which takes about 5 hours. We use a policy head with 3 transformer layers of dimension 1280 and feedforward dimension of 7680. We use the same monotonicity and L2 loss weights from REINA~\cite{reina_hirschkind2025reina} and set $\lambda_{align}=1$. During training we apply VAD trimming randomly to 50\% of our input audios. 
At inference time, we use streaming beam search as implemented in \cite{reina_hirschkind2025reina} of beam size 3  with the same hyperparameters, and simulate streaming using incremental audio chunk sizes of 250ms. We use a temperature of 0.5 for REINA-SAN.

\subsection{Results \& Analysis}
 We first show LAAL vs XCOMET results on combined languages on FLEURS and Europarl in figure~\ref{fig:laal_xcomet_combined}, and select operating points in table~\ref{tab:streaming_results_curated}. REINA-TAN clearly outperforms all other model variants, while REINA-SAN is a close second. The success of REINA-TAN strongly supports our hypothesis that a lack of temporal awareness is the main shortcoming of REINA. Interestingly enough REINA-ALL performs the worst when both methods are combined, suggesting these policies might conflict with each other. We note that despite Whisper-Large V3 having worse S2TT non-streaming performance than Seamless-M4T, we achieve superior streaming efficiency in NoSE scores as shown in table~\ref{tab:nose_fleurs_vs_seamless}.

\subsubsection{Inductive biases of temporal augmentation}
We illustrate the temporal biases of REINA configurations by plotting emission latency against relative token position in figure~\ref{fig:latency-vs-position}. Without the timestep augmented network (TAN), REINA and REINA-SAN exhibit conservative late emissions at the start and end of sentences. Conversely, TAN shifts the model toward a more aggressive, consistent emission schedule to the labels. While the BCE loss in REINA-SAN acts as a rigid global offset that forces earlier emissions that may increase error accumulation, REINA-TAN effectively conditions the policy on elapsed time, allowing for a more stable and efficient emission trajectory throughout the utterance. REINA-ALL suggests that timestep augmentation  and supervised alignment might conflict, and we can see the policy learned in figure~\ref{fig:latency-vs-position} shows a different scheme that does not work as well.

While REINA-TAN provides temporal context that reduces read loops compared to the baseline, BCE loss in REINA-SAN provides a stronger monotonic constraint that effectively eliminates them at high latency. This can be seen with REINA-TAN having 0.024\% Read Loops on FLEURS at 27 BLEU, which is better than REINA having 0.063\% Read Loops on FLEURS at the same BLEU, while REINA-SAN and REINA-ALL do not have these Read Loops at those BLEU values.


\begin{table}[t]
\vspace{-2mm}
\caption{Comparison of Normalized Streaming Efficiency (NoSE) and non-streaming baselines on FLEURS. NoSE values are calculated within the specified latency boundaries $[x, y]$.}
\vspace{-2mm}
\centering
\small
\fontsize{5}{6}\selectfont
\setlength{\tabcolsep}{2pt}
\begin{tabular}{lccc}
\toprule
\textbf{Metric / Model} & \textbf{de$\rightarrow$en} & \textbf{fr$\rightarrow$en} & \textbf{es$\rightarrow$en} \\
\midrule
\textit{NoSE Bounds $[x, y]$ (s)} & $[1.90, 2.05]$ & $[1.67, 1.79]$ & $[1.78, 1.92]$ \\
\midrule
Whisper-Large v3 Non-Streaming (BLEU) & 33.40 & 31.02 & 22.70 \\
SeamlessM4Tv2 (BLEU) & 37.00 & 33.97 & 25.49 \\
\midrule
SeamlessStreaming (NoSE) & 0.925 & 0.940 & 0.936 \\
Whisper-REINA (NoSE) & 0.921 & 0.955 & 0.944 \\
Whisper-REINA-SAN (NoSE) & 0.987 & 0.984 & 0.951 \\
Whisper-REINA-TAN (NoSE) & \textbf{0.991} & \textbf{0.985} & \textbf{0.975} \\
\bottomrule
\end{tabular}

\label{tab:nose_fleurs_vs_seamless}
\vspace{-7mm}
\end{table}

\section{Conclusion}
This paper resolves the temporal drift that limits information-based SimulST policies on large foundation models. We proposed two architectural enhancements: REINA-TAN (continuous temporal embeddings) and REINA-SAN (weak monotonic supervision). The marked success of both methods confirms that a lack of explicit temporal grounding is indeed a major bottleneck causing degenerate waiting states. Evaluated on Whisper Large V3, these enhancements effectively eliminate this issue and establish a new state-of-the-art in streaming efficiency.

\bibliographystyle{IEEEtran}
\bibliography{mybib}

@article{clasi_cheng2024towards,
  title={Towards achieving human parity on end-to-end simultaneous speech translation via llm agent},
  author={Cheng, Shanbo and Huang, Zhichao and Ko, Tom and Li, Hang and Peng, Ningxin and Xu, Lu and Zhang, Qini},
  journal={arXiv preprint arXiv:2407.21646},
  year={2024}
}

@article{hibikizero_labiausse2026simultaneous,
  title={Simultaneous Speech-to-Speech Translation Without Aligned Data},
  author={Labiausse, Tom and Fabre, Romain and Est{\`e}ve, Yannick and D{\'e}fossez, Alexandre and Zeghidour, Neil},
  journal={arXiv preprint arXiv:2602.11072},
  year={2026}
}

@article{reina_hirschkind2025reina,
  title={REINA: Regularized Entropy Information-Based Loss for Efficient Simultaneous Speech Translation},
  author={Hirschkind, Nameer and Liu, Joseph and Yu, Xiao and Nandwana, Mahesh Kumar},
  journal={arXiv preprint arXiv:2508.04946},
  year={2025}
}

@article{hibiki_labiausse2025high,
  title={High-fidelity simultaneous speech-to-speech translation},
  author={Labiausse, Tom and Mazar{\'e}, Laurent and Grave, Edouard and P{\'e}rez, Patrick and D{\'e}fossez, Alexandre and Zeghidour, Neil},
  journal={arXiv preprint arXiv:2502.03382},
  year={2025}
}

@InProceedings{whisper-pmlr-v202-radford23a,
  title = 	 {Robust Speech Recognition via Large-Scale Weak Supervision},
  author =       {Radford, Alec and Kim, Jong Wook and Xu, Tao and Brockman, Greg and Mcleavey, Christine and Sutskever, Ilya},
  booktitle = 	 {Proceedings of the 40th International Conference on Machine Learning},
  pages = 	 {28492--28518},
  year = 	 {2023},
  editor = 	 {Krause, Andreas and Brunskill, Emma and Cho, Kyunghyun and Engelhardt, Barbara and Sabato, Sivan and Scarlett, Jonathan},
  volume = 	 {202},
  series = 	 {Proceedings of Machine Learning Research},
  month = 	 {23--29 Jul},
  publisher =    {PMLR},
  url = 	 {https://proceedings.mlr.press/v202/radford23a.html}
}

@article{seamless_barrault2023seamlessm4t,
  author = {Barrault, Loïc and Chung, Yu-An and Meglioli, Mariano Coria and Dale, David and Dong, Ning and Duquenne, Paul-Ambroise and Elsahar, Hady and Gong, Hongyu and Heffernan, Kevin and Hoffman, John and Klaiber, Christopher and Li, Pengwei and Licht, Daniel and Maillard, Jean and Rakotoarison, Alice and Sadagopan, Kaushik Ram and Wenzek, Guillaume and Ye, Ethan and Akula, Bapi and Chen, Peng-Jen and El Hachem, Naji and Ellis, Brian and Gonzalez, Gabriel Mejia and Haaheim, Justin and Hansanti, Prangthip and Howes, Russ and Huang, Bernie and Hwang, Min-Jae and Inaguma, Hirofumi and Jain, Somya and Kalbassi, Elahe and Kallet, Amanda and Kulikov, Ilia and Lam, Janice and Li, Daniel and Ma, Xutai and Mavlyutov, Ruslan and Peloquin, Benjamin and Ramadan, Mohamed and Ramakrishnan, Abinesh and Sun, Anna and Tran, Kevin and Tran, Tuan and Tufanov, Igor and Vogeti, Vish and Wood, Carleigh and Yang, Yilin and Yu, Bokai and Andrews, Pierre and Balioglu, Can and Costa-jussà, Marta R. and Çelebi, Onur and Elbayad, Maha and Gao, Cynthia and Guzmán, Francisco and Kao, Justine and Lee, Ann and Mourachko, Alexandre and Pino, Juan and Popuri, Sravya and Ropers, Christophe and Saleem, Safiyyah and Schwenk, Holger and Tomasello, Paden and Wang, Changhan and Wang, Jeff and Wang, Skyler and {{SEAMLESS Communication Team}}},
  title = {Joint speech and text machine translation for up to 100 languages},
  journal = {Nature},
  year = {2025},
  volume = {637},
  number = {8046},
  pages = {587--593},
  doi = {10.1038/s41586-024-08359-z},
  url = {https://doi.org/10.1038/s41586-024-08359-z},
  issn = {1476-4687}
}

@inproceedings{waitk-ma-etal-2019-stacl,
    title = "{STACL}: Simultaneous Translation with Implicit Anticipation and Controllable Latency using Prefix-to-Prefix Framework",
    author = "Ma, Mingbo  and
      Huang, Liang  and
      Xiong, Hao  and
      Zheng, Renjie  and
      Liu, Kaibo  and
      Zheng, Baigong  and
      Zhang, Chuanqiang  and
      He, Zhongjun  and
      Liu, Hairong  and
      Li, Xing  and
      Wu, Hua  and
      Wang, Haifeng",
    editor = "Korhonen, Anna  and
      Traum, David  and
      M{\`a}rquez, Llu{\'i}s",
    booktitle = "Proceedings of the 57th Annual Meeting of the Association for Computational Linguistics",
    month = jul,
    year = "2019",
    address = "Florence, Italy",
    publisher = "Association for Computational Linguistics",
    url = "https://aclanthology.org/P19-1289/",
    doi = "10.18653/v1/P19-1289",
    pages = "3025--3036"
}

@inproceedings{ma2020monotonic,
    title = "{S}imul{MT} to {S}imul{ST}: Adapting Simultaneous Text Translation to End-to-End Simultaneous Speech Translation",
    author = "Ma, Xutai  and
      Pino, Juan  and
      Koehn, Philipp",
    editor = "Wong, Kam-Fai  and
      Knight, Kevin  and
      Wu, Hua",
    booktitle = "Proceedings of the 1st Conference of the Asia-Pacific Chapter of the Association for Computational Linguistics and the 10th International Joint Conference on Natural Language Processing",
    month = dec,
    year = "2020",
    address = "Suzhou, China",
    publisher = "Association for Computational Linguistics",
    url = "https://aclanthology.org/2020.aacl-main.58/",
    doi = "10.18653/v1/2020.aacl-main.58",
    pages = "582--587"
}

@inproceedings{whisperx_Bain2023WhisperXTS,
  title={WhisperX: Time-Accurate Speech Transcription of Long-Form Audio},
  author={Max Bain and Jaesung Huh and Tengda Han and Andrew Zisserman},
  booktitle={Interspeech},
  year={2023},
  url={https://api.semanticscholar.org/CorpusID:257255343}
}

@inproceedings{arivazhagan2019monotonic,
    title = "Monotonic Infinite Lookback Attention for Simultaneous Machine Translation",
    author = "Arivazhagan, Naveen  and
      Cherry, Colin  and
      Macherey, Wolfgang  and
      Chiu, Chung-Cheng  and
      Yavuz, Semih  and
      Pang, Ruoming  and
      Li, Wei  and
      Raffel, Colin",
    editor = "Korhonen, Anna  and
      Traum, David  and
      M{\`a}rquez, Llu{\'i}s",
    booktitle = "Proceedings of the 57th Annual Meeting of the Association for Computational Linguistics",
    month = jul,
    year = "2019",
    address = "Florence, Italy",
    publisher = "Association for Computational Linguistics",
    url = "https://aclanthology.org/P19-1126/",
    doi = "10.18653/v1/P19-1126",
    pages = "1313--1323"
}

@article{graves2012sequence,
  author       = {Alex Graves},
  title        = {Sequence Transduction with Recurrent Neural Networks},
  journal      = {CoRR},
  volume       = {abs/1211.3711},
  year         = {2012},
  url          = {http://arxiv.org/abs/1211.3711},
  eprinttype    = {arXiv},
  eprint       = {1211.3711},
  bibsource    = {dblp computer science bibliography, https://dblp.org}
}

@inproceedings{xue2022large,
  title={Large-Scale Streaming End-to-End Speech Translation with Neural Transducers},
  author={Jian Xue and Peidong Wang and Jinyu Li and Matt Post and Yashesh Gaur},
  booktitle={Interspeech},
  year={2022},
  url={https://api.semanticscholar.org/CorpusID:248118691}
}

@inproceedings{canary,
author = {Puvvada, Krishna and Żelasko, Piotr and Huang, He and Hrinchuk, Oleksii and Koluguri, Nithin and Dhawan, Kunal and Majumdar, Somshubra and Rastorgueva, Elena and Chen, Zhehuai and Lavrukhin, Vitaly and Balam, Jagadeesh and Ginsburg, Boris},
year = {2024},
month = {09},
pages = {3964-3968},
title = {Less is More: Accurate Speech Recognition \& Translation without Web-Scale Data},
doi = {10.21437/Interspeech.2024-2294}
}

@inproceedings{owsm,
author = {Peng, Yifan and Tian, Jinchuan and Chen, William and Arora, Siddhant and Yan, Brian and Sudo, Yui and Shakeel, Muhammad and Choi, Kwanghee and Shi, Jiatong and Chang, Xuankai and Jung, Jee-Weon and Watanabe, Shinji},
year = {2024},
month = {09},
pages = {352-356},
title = {OWSM v3.1: Better and Faster Open Whisper-Style Speech Models based on E-Branchformer},
doi = {10.21437/Interspeech.2024-1194}
}

@inproceedings{ren-etal-2020-simulspeech,
    title = "{S}imul{S}peech: End-to-End Simultaneous Speech to Text Translation",
    author = "Ren, Yi  and
      Liu, Jinglin  and
      Tan, Xu  and
      Zhang, Chen  and
      Qin, Tao  and
      Zhao, Zhou  and
      Liu, Tie-Yan",
    editor = "Jurafsky, Dan  and
      Chai, Joyce  and
      Schluter, Natalie  and
      Tetreault, Joel",
    booktitle = "Proceedings of the 58th Annual Meeting of the Association for Computational Linguistics",
    month = jul,
    year = "2020",
    address = "Online",
    publisher = "Association for Computational Linguistics",
    url = "https://aclanthology.org/2020.acl-main.350/",
    doi = "10.18653/v1/2020.acl-main.350",
    pages = "3787--3796"
}

@inproceedings{liu2020low,
author = {Liu, Danni and Spanakis, Gerasimos and Niehues, Jan},
year = {2020},
month = {10},
pages = {3620-3624},
title = {Low-Latency Sequence-to-Sequence Speech Recognition and Translation by Partial Hypothesis Selection},
doi = {10.21437/Interspeech.2020-2897}
}

@article{dig_sst_2024, 
    title={Divergence-Guided Simultaneous Speech Translation},
    volume={38},
    url={https://ojs.aaai.org/index.php/AAAI/article/view/29733},
    DOI={10.1609/aaai.v38i16.29733},
    abstractNote={To achieve high-quality translation with low latency, a Simultaneous Speech Translation (SimulST) system relies on a policy module to decide whether to translate immediately or wait for additional streaming input, along with a translation model capable of effectively handling partial speech input. Prior research has tackled these components separately, either using ``wait-k’’ policies based on fixed-length segments or detected word boundaries, or dynamic policies based on different strategies (e.g., meaningful units), while employing offline models for prefix-to-prefix translation. In this paper, we propose Divergence-Guided Simultaneous Speech Translation (DiG-SST), a tightly integrated approach focusing on both translation quality and latency for streaming input. Specifically, we introduce a simple yet effective prefix-based strategy for training translation models with partial speech input, and develop an adaptive policy that makes read/write decisions for the translation model based on the expected divergence in translation distributions resulting from future input. Our experiments on multiple translation directions of the MuST-C benchmark demonstrate that our approach achieves a better trade-off between translation quality and latency compared to existing methods.},
    number={16},
    journal={Proceedings of the AAAI Conference on Artificial Intelligence},
    author="Chen, Xinjie and Fan, Kai and Luo, Wei and Zhang, Linlin and Zhao, Libo and Liu, Xinggao and Huang, Zhongqiang", 
    year={2024}, 
    month={Mar.}, 
    pages={17799-17807} 
}

@article{vaswani2017attention,
  title={Attention is all you need},
  author={Vaswani, Ashish and Shazeer, Noam and Parmar, Niki and Uszkoreit, Jakob and Jones, Llion and Gomez, Aidan N and Kaiser, {\L}ukasz and Polosukhin, Illia},
  journal={Advances in neural information processing systems},
  volume={30},
  year={2017}
}

@INPROCEEDINGS{fleurs,
  author={Conneau, Alexis and Ma, Min and Khanuja, Simran and Zhang, Yu and Axelrod, Vera and Dalmia, Siddharth and Riesa, Jason and Rivera, Clara and Bapna, Ankur},
  booktitle={2022 IEEE Spoken Language Technology Workshop (SLT)}, 
  title={FLEURS: FEW-Shot Learning Evaluation of Universal Representations of Speech}, 
  year={2023},
  volume={},
  number={},
  pages={798-805},
  doi={10.1109/SLT54892.2023.10023141}
}

@inproceedings{europarl,
    title = "{E}uroparl: A Parallel Corpus for Statistical Machine Translation",
    author = "Koehn, Philipp",
    booktitle = "Proceedings of Machine Translation Summit X: Papers",
    month = sep # " 13-15",
    year = "2005",
    address = "Phuket, Thailand",
    url = "https://aclanthology.org/2005.mtsummit-papers.11/",
    pages = "79--86"
}

@unknown{easist,
author = {Fu, Biao and Yu, Donglei and Liao, Minpeng and Li, Chengxi and Chen, Yidong and Fan, Kai},
year = {2025},
month = {04},
pages = {},
title = {Efficient and Adaptive Simultaneous Speech Translation with Fully Unidirectional Architecture},
doi = {10.48550/arXiv.2504.11809}
}

@inproceedings{metricx,
    title = "{M}etric{X}-24: The {G}oogle Submission to the {WMT} 2024 Metrics Shared Task",
    author = "Juraska, Juraj  and
      Deutsch, Daniel  and
      Finkelstein, Mara  and
      Freitag, Markus",
    editor = "Haddow, Barry  and
      Kocmi, Tom  and
      Koehn, Philipp  and
      Monz, Christof",
    booktitle = "Proceedings of the Ninth Conference on Machine Translation",
    month = nov,
    year = "2024",
    address = "Miami, Florida, USA",
    publisher = "Association for Computational Linguistics",
    url = "https://aclanthology.org/2024.wmt-1.35/",
    doi = "10.18653/v1/2024.wmt-1.35",
    pages = "492--504"
}

@article{team2024gemma,
  title={Gemma 2: Improving open language models at a practical size},
  author={Team, Gemma and Riviere, Morgane and Pathak, Shreya and Sessa, Pier Giuseppe and Hardin, Cassidy and Bhupatiraju, Surya and Hussenot, L{\'e}onard and Mesnard, Thomas and Shahriari, Bobak and Ram{\'e}, Alexandre and others},
  journal={arXiv preprint arXiv:2408.00118},
  year={2024}
}

@inproceedings{cvss,
  title={CVSS Corpus and Massively Multilingual Speech-to-Speech Translation},
  author={Jia, Ye and Ramanovich, Michelle Tadmor and Wang, Quan and Zen, Heiga},
  booktitle={Proceedings of the Thirteenth Language Resources and Evaluation Conference},
  pages={6691--6703},
  year={2022}
}

@inproceedings{pratap2020mls,
  title={MLS: A Large-Scale Multilingual Dataset for Speech Research},
  author={Pratap, Vineel and Xu, Qiantong and Sriram, Anuroop and Synnaeve, Gabriel and Collobert, Ronan},
  booktitle={Proc. Interspeech 2020},
  pages={2757--2761},
  year={2020}
}

@misc{qwen3,
      title={Qwen3 Technical Report}, 
      author={An Yang and Anfeng Li and Baosong Yang and Beichen Zhang and Binyuan Hui and Bo Zheng and Bowen Yu and Chang Gao and Chengen Huang and Chenxu Lv and Chujie Zheng and Dayiheng Liu and Fan Zhou and Fei Huang and Feng Hu and Hao Ge and Haoran Wei and Huan Lin and Jialong Tang and Jian Yang and Jianhong Tu and Jianwei Zhang and Jianxin Yang and Jiaxi Yang and Jing Zhou and Jingren Zhou and Junyang Lin and Kai Dang and Keqin Bao and Kexin Yang and Le Yu and Lianghao Deng and Mei Li and Mingfeng Xue and Mingze Li and Pei Zhang and Peng Wang and Qin Zhu and Rui Men and Ruize Gao and Shixuan Liu and Shuang Luo and Tianhao Li and Tianyi Tang and Wenbiao Yin and Xingzhang Ren and Xinyu Wang and Xinyu Zhang and Xuancheng Ren and Yang Fan and Yang Su and Yichang Zhang and Yinger Zhang and Yu Wan and Yuqiong Liu and Zekun Wang and Zeyu Cui and Zhenru Zhang and Zhipeng Zhou and Zihan Qiu},
      year={2025},
      eprint={2505.09388},
      archivePrefix={arXiv},
      primaryClass={cs.CL},
      url={https://arxiv.org/abs/2505.09388}, 
}

@inproceedings{laal,
    title = "Over-Generation Cannot Be Rewarded: Length-Adaptive Average Lagging for Simultaneous Speech Translation",
    author = "Papi, Sara  and
      Gaido, Marco  and
      Negri, Matteo  and
      Turchi, Marco",
    editor = "Ive, Julia  and
      Zhang, Ruiqing",
    booktitle = "Proceedings of the Third Workshop on Automatic Simultaneous Translation",
    month = jul,
    year = "2022",
    address = "Online",
    publisher = "Association for Computational Linguistics",
    url = "https://aclanthology.org/2022.autosimtrans-1.2/",
    doi = "10.18653/v1/2022.autosimtrans-1.2",
    pages = "12--17"
}

@inproceedings{sacrebleu,
    title = "A Call for Clarity in Reporting {BLEU} Scores",
    author = "Post, Matt",
    editor = "Bojar, Ond{\v{r}}ej  and
      Chatterjee, Rajen  and
      Federmann, Christian  and
      Fishel, Mark  and
      Graham, Yvette  and
      Haddow, Barry  and
      Huck, Matthias  and
      Yepes, Antonio Jimeno  and
      Koehn, Philipp  and
      Monz, Christof  and
      Negri, Matteo  and
      N{\'e}v{\'e}ol, Aur{\'e}lie  and
      Neves, Mariana  and
      Post, Matt  and
      Specia, Lucia  and
      Turchi, Marco  and
      Verspoor, Karin",
    booktitle = "Proceedings of the Third Conference on Machine Translation: Research Papers",
    month = oct,
    year = "2018",
    address = "Brussels, Belgium",
    publisher = "Association for Computational Linguistics",
    url = "https://aclanthology.org/W18-6319/",
    doi = "10.18653/v1/W18-6319",
    pages = "186--191"
}

@article{xcomet,
    title = "x{COMET}: Transparent Machine Translation Evaluation through Fine-grained Error Detection",
    author = "Guerreiro, Nuno M.  and
      Rei, Ricardo  and
      Stigt, Daan van  and
      Coheur, Luisa  and
      Colombo, Pierre  and
      Martins, Andr{\'e} F. T.",
    journal = "Transactions of the Association for Computational Linguistics",
    volume = "12",
    year = "2024",
    address = "Cambridge, MA",
    publisher = "MIT Press",
    url = "https://aclanthology.org/2024.tacl-1.54/",
    doi = "10.1162/tacl_a_00683",
    pages = "979--995"
}

@misc{emma,
      title={Efficient Monotonic Multihead Attention}, 
      author={Xutai Ma and Anna Sun and Siqi Ouyang and Hirofumi Inaguma and Paden Tomasello},
      year={2023},
      eprint={2312.04515},
      archivePrefix={arXiv},
      primaryClass={cs.CL},
      url={https://arxiv.org/abs/2312.04515}, 
}

@article{sasst,
  title={SASST: Leveraging Syntax-Aware Chunking and LLMs for Simultaneous Speech Translation},
  author={Yang, Zeyu and Wei, Lai and Koshkin, Roman and Chen, Xi and Nakamura, Satoshi},
  journal={arXiv preprint arXiv:2508.07781},
  year={2025}
}

@article{seedliveinterpret_cheng2025seed,
  title={Seed LiveInterpret 2.0: End-to-end Simultaneous Speech-to-speech Translation with Your Voice},
  author={Cheng, Shanbo and Bao, Yu and Huang, Zhichao and Lu, Yu and Peng, Ningxin and Xu, Lu and Yu, Runsheng and Cao, Rong and Du, Yujiao and Han, Ting and others},
  journal={arXiv preprint arXiv:2507.17527},
  year={2025}
}

\end{document}